\documentclass[sigconf]{acmart}
\AtBeginDocument{%
  }
    
\usepackage{makecell}
\usepackage{tabularx}
\usepackage{multirow}
\usepackage{balance} 
\copyrightyear{2025}
\acmYear{2025}
\setcopyright{acmlicensed}
\acmConference[MM '25] {Proceedings of the 33rd ACM International Conference on Multimedia}{October 27--31, 2025}{Dublin, Ireland.}
\acmBooktitle{Proceedings of the 33rd ACM International Conference on Multimedia (MM '25), October 27--31, 2025, Dublin, Ireland}
\acmISBN{979-8-4007-2035-2/2025/10}
\acmDOI{10.1145/3746027.3754943}

\begin{document}

\title{LaVieID: Local Autoregressive Diffusion Transformers for Identity-Preserving Video Creation}


\author{Wenhui Song}
\authornote{Both authors contributed equally to this research.}
\affiliation{%
  \institution{Shenzhen Campus of Sun Yat-sen University}
  \city{Shenzhen}
  \country{China}}
\email{wenhuisong16@gmail.com}

\author{Hanhui Li}
\authornotemark[1]
\affiliation{%
  \institution{Shenzhen Campus of Sun Yat-sen University}
  \city{Shenzhen}
  \country{China}
}
\email{lihh77@mail.sysu.edu.cn}

\author{Jiehui Huang}
\affiliation{%
 \institution{Hong Kong University of Science and Technology (HKUST)}
 \city{Hong Kong}
 \country{China}}
\email{jhhuang117@gmail.com}

\author{Panwen Hu}
\affiliation{%
  \institution{Mohamed bin Zayed University of Artificial Intelligence}
  \city{Abu Dhabi}
  \country{United Arab Emirates}}
\email{Panwen.Hu@mbzuai.ac.ae}

\author{Yuhao Cheng}
\affiliation{%
  \institution{Lenovo Research}
  \city{Beijing}
  \country{China}}
\email{chengyh5@lenovo.com}

\author{Long Chen}
\affiliation{%
  \institution{Lenovo Research}
  \city{Beijing}
  \country{China}}
\email{chenlong12@lenovo.com}

\author{Yiqiang Yan}
\affiliation{%
  \institution{Lenovo Research}
  \city{Beijing}
  \country{China}}
\email{yanyq@lenovo.com}

\author{Xiaodan Liang}
\authornote{Corresponding author.}
\affiliation{%
  \institution{Shenzhen Campus of Sun Yat-sen University}
  \city{Shenzhen}
  \country{China}}
\email{liangxd9@mail.sysu.edu.cn}

\renewcommand{\shortauthors}{Wenhui Song et al.}

\begin{abstract}
In this paper, we present LaVieID, a novel \underline{l}ocal \underline{a}utoregressive \underline{vi}d\underline{e}o diffusion framework designed to tackle the challenging \underline{id}entity-preserving text-to-video task. The key idea of LaVieID is to mitigate the loss of identity information inherent in the stochastic global generation process of diffusion transformers (DiTs) from both spatial and temporal perspectives. Specifically, unlike the global and unstructured modeling of facial latent states in existing DiTs, LaVieID introduces a local router to explicitly represent latent states by weighted combinations of fine-grained local facial structures. This alleviates undesirable feature interference and encourages DiTs to capture distinctive facial characteristics. Furthermore, a temporal autoregressive module is integrated into LaVieID to refine denoised latent tokens before video decoding. This module divides latent tokens temporally into chunks, exploiting their long-range temporal dependencies to predict biases for rectifying tokens, thereby significantly enhancing inter-frame identity consistency. Consequently, LaVieID can generate high-fidelity personalized videos and achieve state-of-the-art performance. Our code and models are available at \url{https://github.com/ssugarwh/LaVieID}.

\end{abstract}



\maketitle

\section{Introduction}
\label{sec:intro}
In recent years, diffusion models \cite{croitoru2023diffusion} have demonstrated remarkable capabilities in modeling data of various modalities, such as texts \cite{gongdiffuseq}, time series \cite{tashiro2021csdi}, images \cite{saharia2022palette}, 3D models \cite{xie2024dreamvton} and videos \cite{ho2022video}. These advances have significantly contributed to the development and application of content synthesis, especially for tasks such as text-to-image (T2I) \cite{ramesh2021zero} and text-to-video (T2V) \cite{singermake2023} generation. However, the emerging task of identity-preserving T2V (IPT2V) generation conditioned on a single subject image \cite{he2024id,chefer2024still,yuan2024identity}, remains challenging for existing generative models. This is because it needs to explore large latent spaces and requires higher spatio-temporal consistency, compared with personalized T2I generation \cite{guo2024pulid} or subject animation with reference motions \cite{wei2024dreamvideo,huang2024magicfight}.  


\begin{figure*}
      \includegraphics[width=\textwidth]{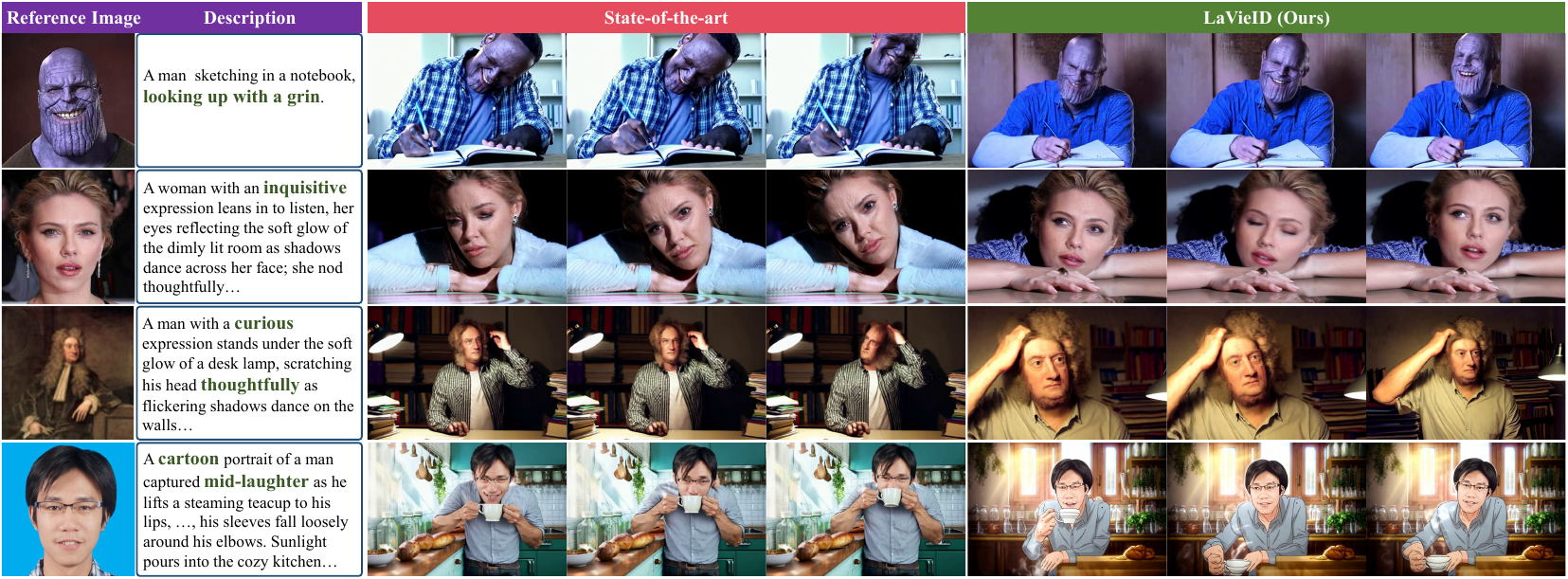}
  \caption{Given a single reference image of a target person, the proposed LaVieID framework can follow diverse instructions to generate vivid videos with identity fidelity and visual quality superior to that of state-of-the-art methods \cite{yuan2024identity}.}
  \label{fig:teaser}
\end{figure*}

Several pioneering studies seek to address IPT2V generation by leveraging cutting-edge diffusion transformers (DiTs) \cite{peebles2023scalable} trained on large-scale data, and enhancing them with carefully designed modules and strategies, such as textual inversion \cite{ma2024magic}, frequency decomposition \cite{yuan2024identity}, cross-video pairing \cite{zhong2025concat}, and reinforcement learning \cite{li2025magicid}. Although these methods improve spatio-temporal consistency to some extent, they still suffer from the loss of identity information due to the intrinsic architecture designs of DiTs. Particularly, current DiTs \cite{yang2024cogvideox} primarily rely on global attention modules to capture spatio-temporal feature correlations, which are relatively insensitive to fine-grained facial structures and temporal dependencies. Hence, there is a substantial risk that identity-related features may be disrupted by irrelevant information, yielding issues like identity incoherence, abrupt motions, and flickering frames. 

Therefore, in order to address the aforementioned challenges, this paper proposes LaVieID, a local autoregressive video diffusion framework to tackle the above issues. LaVieID improves the identity preservation capabilities of DiTs from both spatial and temporal perspectives. First, LaVieID introduces a local router that leverages fine-grained facial structures, such as eyebrows, eyes, and mouth, to provide spatial guidance on modeling feature correlations. Specifically, the proposed local router learns to reconstruct and refine the latent tokens of subjects based on local facial structures, with adaptive weights that reflect the relative importance of these structures. In this way, latent states related to facial structures and subject identity are emphasized, while those that are irrelevant are suppressed, and consequently mitigate feature interference. Furthermore, a temporal autoregressive module is introduced to enhance temporal identity consistency. This module splits the denoised latent tokens of videos into multiple sequential chunks and explicitly establishes correlations between adjacent chunks, thus improving long-range temporal dependencies across frames. With the local router and the temporal autoregressive module, our LaVieID framework effectively overcomes the limitations of existing diffusion models in identity preserving, and achieves state-of-the-art performance in IPT2V generation, as shown in Figure \ref{fig:teaser}.

In summary, LaVieID provides conventional DiTs with global and unstructured attention with the ability to exploit local facial structures and model long-range temporal dependencies, facilitating video customization with enhanced spatio-temporal identity consistency and visual quality. The main contributions of this paper are summarized as follows:

$\bullet$ We propose LaVieID, a novel framework that effectively addresses the identity-preserving text-to-video generation task.

$\bullet$ We introduce a local router to provide spatial structural guidance by leveraging fine-grained facial cues. 

$\bullet$ We devise a temporal autoregressive module to model long-range temporal dependencies among frames.  

$\bullet$ Comprehensive quantitative and qualitative analyses and a user study validate that LaVieID outperforms state-of-the-art methods.

\section{Related Work}
\label{sec:related_work}

\subsection{Generative Models}
\textbf{Diffusion models.} Diffusion-based video generation models can be broadly categorized into UNet-based and transformer-based approaches. Among \emph{UNet-based methods}, an early approach \cite{ho2022video} extends the UNet architecture along the temporal dimension (using a dilated 3D UNet) to adapt it for video generation. Building on this, the latent video diffusion model \cite{he2022latent} shifts from pixel space to latent space to represent video frames by a lower dimension. This effectively reduces computational and memory requirements, albeit at the cost of certain fine details. Recently, \emph{transformer-based video diffusion models} have attracted considerable attention due to their scalability. Notably, Latte \cite{ma2024latte} introduces one of the first T2V diffusion models using a DiT architecture, which paves the way for a series of subsequent models, such as Sora \cite{openai2023videomodel}, Open-Sora \cite{PKU2024OpenSoraPlan}, and CogVideoX \cite{yang2024cogvideox}. These models have achieved impressive progress in simulating the physical world. However, they still face severe challenges in maintaining identity consistency across frames. 

\textbf{Autoregressive models.} The dominant paradigm of autoregression models is \emph{Next-token prediction} that sequentially generates discrete tokens obtained via vector quantization methods \cite{van2017neural}. Representative autoregression-based models include CogVideo \cite{hong2022cogvideo} and VideoPoet \cite{kondratyuk2023videopoet}, and both of them employ transformer-based architectures to generate video tokens autoregressively conditioned on previous context tokens. Specifically, CogVideo integrates hierarchical and multi-frame-rate training to enhance temporal coherence aligned with textual prompts, while VideoPoet leverages multi-modal inputs for flexible video synthesis. MAGVIT \cite{yu2023magvit} employs masked token prediction techniques to accelerate inference speeds, and NOVA \cite{deng2024autoregressive} uses continuous-valued tokens and intra-frame bidirectional modeling for efficient generation.  Teller \cite{zhen2025teller} introduces an autoregressive motion generation framework for real-time streaming portrait animation, driven by audio inputs. It leverages causal temporal attention and motion prediction priors to ensure responsiveness and consistency under streaming conditions. Meanwhile, Neighboring Autoregressive Modeling\cite{he2025neighboring} proposes a local temporal conditioning scheme that only attends to nearby frame tokens, significantly improving short-term motion realism and reducing autoregressive redundancy in long sequences.However, the discreteness of tokens inherently causes spatio-temporal incoherence. More recently, a new \emph{next-scale prediction} paradigm \cite{tian2024visual,li2024controlvar} tries to handle this issue via predicting multi-scale latent representations of images, yet its employment to video generation remains unexplored. Several studies \cite{ge2022long,han2022show,ren2025next,gu2025long} design strategies for improving temporal content consistency in the paradigm of frame interpolation or extrapolation. Nevertheless, the general performance of autoregression-based methods still requires further improvement to rival that of diffusion models.

\textbf{Hybrid models.} Hybrid models integrate autoregression methods with diffusion models to exploit the temporal consistency of autoregression approaches alongside the high-quality generation capabilities of diffusion models. For example, ACDiT \cite{hu2024acdit} employs a block-wise conditional diffusion mechanism within an autoregression framework, preserving temporal coherence across video segments. MaskFlow \cite{fuest2025maskflow} introduces discrete token-based flow matching within an autoregression setting, significantly accelerating video generation and ensuring frame continuity. CausVid \cite{yin2024slow} applies autoregression diffusion distillation, considerably reducing diffusion inference steps and enabling near real-time generation without compromising visual quality or temporal coherence. VideoWorld \cite{ren2025videoworld} integrates interactive prompting into an autoregression-diffusion hybrid framework, facilitating user-guided dynamic video generation. AR-Diffusion \cite{sun2025ar} introduces an asynchronous latent diffusion framework conditioned on past latent features in a chunk-wise autoregressive manner. By decoupling motion prediction from high-fidelity synthesis, it enables temporally coherent and efficient video generation with fewer sampling steps.Yet these methods are complex and inevitably modify the original latent space of diffusion models, which requires expensive computational costs and training data. Moreover, they still lack structured modeling strategies of spatio-temporal identity consistency, which is essential for the IPT2V task.

\begin{figure*}[!t]
\centering
\includegraphics[width=\linewidth]{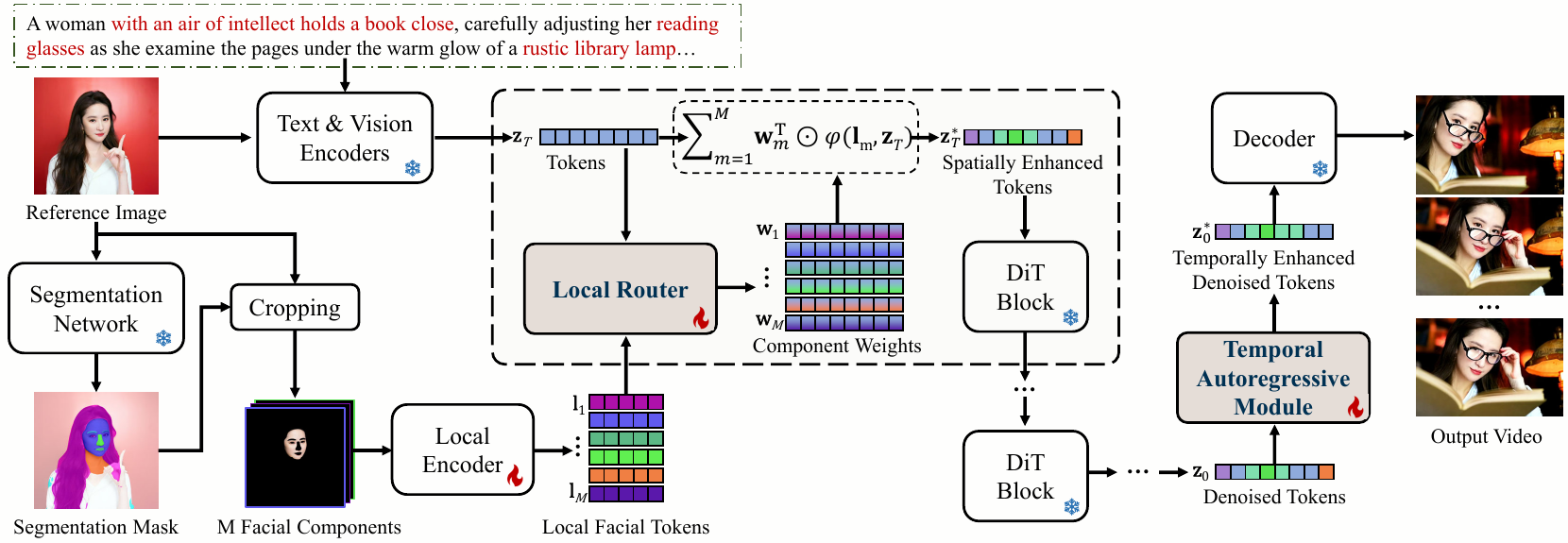}
\caption{The proposed LaVieID framework. Two key components are introduced to improve the baseline DiT in spatio-temporal identity preservation: (i) A local router exploits local facial tokens to incorporate spatial structures into latent tokens; (ii) A temporal autoregressive module establishes long-range temporal dependencies in denoised tokens.}
\label{fig:framework}
\end{figure*}

\subsection{Personalized Content Synthesis}
\textbf{Personalized image synthesis.} Recent advancements in image synthesis have significantly progressed towards producing high-quality, identity-consistent visuals. OneDiffusion \cite{le2024one} formulates personalized synthesis tasks within a unified sequential modeling framework, enabling tasks such as inpainting and novel view synthesis by treating various conditional images as diverse viewpoints of a single target. To further enhance the quality and personalization, InfiniteYou \cite{jiang2025infiniteyou} introduces a multi-stage training approach incorporating supervised fine-tuning on synthesized personalized images, thus notably improving visual fidelity and identity consistency. ConsistentID \cite{huang2024consistentid} exploits local facial components to preserve identity information by inserting their visual features into text tokens. This strategy is effective in identity preservation, but it also yields text contamination and is insufficient for feature disentanglement. To address contamination of identity features, PuLID \cite{guo2024pulid} leverages contrastive alignment to facilitate clean insertion of identity embeddings, mitigating undesirable feature blending.  

\textbf{Personalized video synthesis.} Personalized video synthesis emphasizes both spatial and temporal identity consistency, thus facing additional challenges compared with personalized image synthesis. ConsisID \cite{yuan2024identity} proposes to decompose facial information into low-frequency and high-frequency components, which are modeled by a keypoint-based global extractor and a local extractor supervised by face recognition, respectively. ID-Animator \cite{he2024id} addresses identity preservation in a zero-shot manner by leveraging latent facial queries and identity-oriented datasets without requiring task-specific fine-tuning. PersonVideo \cite{li2024personalvideo} uses multi-level personalized tokens to maintain identity across video sequences while allowing precise text following. Concat-ID \cite{zhong2025concat} introduces a scalable framework that integrates variational autoencoders with video latents through 3D self-attention, leveraging cross-video pairing and multi-stage training to tackle diverse and complex video scenarios. Magic Mirror \cite{zhang2025magic} integrates dual-branch facial feature extraction with conditioned adaptive normalization within a DiT, effectively balancing identity consistency and dynamic motion generation. Similarly, FantasyID \cite{zhang2025fantasyid} incorporates explicit 3D facial geometry priors and multi-view augmentation into diffusion transformers, enhancing the structural stability of facial identities. In parallel, approaches oriented toward animation tasks, such as AniPortrait \cite{wei2024aniportrait}, FaceShot \cite{gaofaceshot}, Hallo2 \cite{cui2024hallo2}, and LivePortrait \cite{guo2024liveportrait}, have focused primarily on facial motion synchronization or expression control. Although these aforementioned methods substantially enhance the flexibility and quality of video generation, the global and unstructured feature correlations in their architectures still hinder their abilities in identity preservation.

\section{Methodology}
\label{sec:method}
We present the details of LaVieID in this section. We begin by formulating the identity-preserving text-to-video (IPT2V) generation task and providing the overall framework of LaVieID in Sec. \ref{sec:framework}. We then introduce the two key components of LaVieID, including a local router (Sec. \ref{sec:router}) and a temporal autoregressive module (Sec. \ref{sec:ar}) for enhancing spatial and temporal identity consistency, respectively. At last, the learning objectives of LaVieID are provided in Sec. \ref{sec:loss}.

\subsection{Overall Framework}
\label{sec:framework}

\textbf{Task formulation.} Given a reference image of a target person and a text prompt, our goal is to generate a corresponding video of the target person with spatially and temporally coherent identity. Formally, we formulate the video generation process by a latent diffusion model (LDM)~\cite{ho2020denoising,yang2024cogvideox} guided by the following learning objective: 
\begin{equation}
\mathcal{L}_{diff}=\mathbb{E}_{t,\epsilon,\mathbf{z}_0}||\epsilon-\epsilon_{\theta}(\sqrt{\bar{a}_t}\mathbf{z}_0 + \sqrt{1-\bar{a}_t}\epsilon,t)||^{2},
\label{eq:l_diff}
\end{equation}
where $\epsilon_{\theta}$ denotes the LDM with trainable parameters $\theta$. $\epsilon_{\theta}$ can be interpreted as an iterative process consisting of $T$ denoising steps that gradually restores the joint latent representation of the reference image and the text prompt $\mathbf{z}_0$ from a random noise $\epsilon \sim \mathcal{N}(\mathbf{0},\mathbf{I})$. $t \in[1, T]$ denotes an arbitrary denoising step and $\bar{a}_t$ is a hyperparameter controlling the noise scale at the $t$-th step. 

In this paper, we implement $\epsilon_{\theta}$ based on a cutting-edge text-to-video model \cite{yang2024cogvideox}, in which multiple DiT blocks with 3D hybrid full attention are used to model global spatio-temporal contexts. However, these attention modules disregard facial structures and calculate correlations across all latent tokens, which are prone to feature interference and loss of target identity. Hence, we develop the following LaVieID framework to address these limitations.

\textbf{LaVieID framework.} The framework of LaVieID is shown in Figure \ref{fig:framework}. In parallel with the text and vision encoders of the baseline model, an off-the-shelf segmentation network \cite{yu2018bisenet} is employed to extract $M$ local facial components (such as eyes, nose, and lips) from the target image. Subsequently, a local encoder embeds these facial components into $M$ corresponding token sequences. As the subject structures in synthesized videos are mainly influenced by front DiT blocks \cite{chen2024delta}, we propose to inject the facial token sequences into the first DiT block of the baseline model via a local router. This router is designed to model the correlations between the joint latent tokens and the local facial token sequences, so that the facial structures can be leveraged adaptively to refine the joint latent tokens and better preserve identity information. The latent tokens processed by all DiT blocks after $T$ iterations are noise-free. Before decoding them into the output video, we further improve their temporal consistency via an autoregressive module. This module splits the denoised tokens into multiple chunks temporally (i.e., along the frame axis), and progressively refines the tokens in one chunk by predicting their biases conditioned on those in the preceding chunk. This ensures explicit long-range temporal dependencies among the denoised tokens, thus overcoming the limitations of the temporally independent and unstructured denoising process of the baseline model. Consequently, the proposed LaVieID framework is capable of generating latent representations with high spatio-temporal identity consistency to address the IPT2V task.

\begin{figure*}[!t]
\centering
\includegraphics[width=\linewidth]{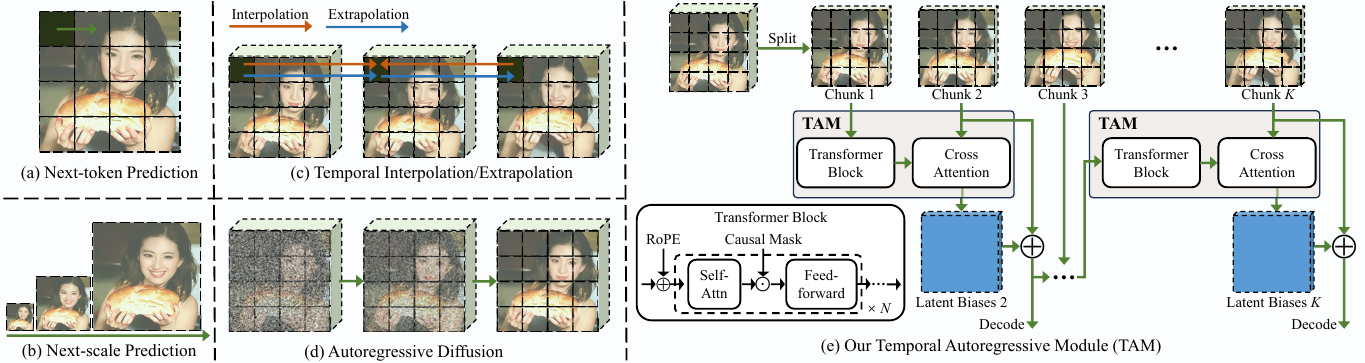}
\caption{The proposed temporal autoregressive module against other autoregression-based mechanisms. Existing autoregression methods rely on predicting (a) discrete tokens \cite{yulanguage,kondratyuk2024videopoet}, (b) multi-scale tokens \cite{tian2024visual,li2024controlvar}, or (c) inter-frame tokens \cite{ge2022long,han2022show}, of which the visual quality remains inferior to that of diffusion models. (d) Hybrid models \cite{ho2022video,chen2024diffusion,ma2024janusflow,ruhe2024rolling} leverage the advantages of diffusion models in visual quality and autoregression models in temporal modeling. Yet their schemes are complicated and require intensive resources. (e) Our module models long-range temporal dependencies across frames on the chunk-wise level. Note that it is applied on denoised tokens, and hence it will not alter the latent space of the baseline DiT while enjoying the advantages of both diffusion models and autoregression models efficiently.}
\label{fig:tam}
\end{figure*}

\subsection{Local Router}
\label{sec:router}
The proposed local router is devised to disentangle global facial information into fine-grained local representations, thereby enhancing spatial identity consistency. The pioneering research on personalized image synthesis \cite{huang2024consistentid} also proposes to exploit local facial components. Nevertheless, it simply incorporates component-level features into text tokens, which contaminates textual information and hinders the ability of its model in feature disentanglement. In contrast, our local router estimates and modulates the influences of local facial components on the joint latent tokens, thus it can emphasize identity-related tokens while suppressing those that are irrelevant to better preserve identity information.        

Specifically, let $\mathbf{l}_m \in \mathbb{R}^{L\times D}$ denote one of the local facial token sequences, where $L$ is the number of tokens, $D$ is the feature dimension, and $ m\in[1, M]$. For the conciseness of presentation, we omit the index of denoising time step and denote the joint latent tokens as $\mathbf{z} \in \mathbb{R}^{L' \times D'}$. The proposed local router calculates the token-wise weights for each component ${{\bf{w}}_m} \in [0, 1]^{1\times L'}$ as follows:       
\begin{equation}
    {{\bf{w}}_m} = \sigma {({{\bf{W}}_m}\tilde{{\bf{l}}_m}{{\bf{W}}_l}{\bf{W}}_z^{\rm{T}}{\tilde{\bf{z}}}^{\rm{T}})},
    \label{eq:weight}
\end{equation}
where $\tilde{{\bf{l}}_m}$ and $\tilde{\bf{z}}$ are $\mathbf{l}_m$ and $\mathbf{z}$ after layer normalization \cite{ba2016layer}. ${\bf{W}}_l \in \mathbb{R}^{D\times D''}$ and ${\bf{W}}_z \in \mathbb{R}^{D'\times D''}$ are two linear transformations that project ${\bf{l}}_m$ and $\bf{z}$ into the same latent space. $\rm{T}$ represents the matrix transpose operator. It is easy to see that ${\tilde{{\bf{l}}_m}{{\bf{W}}_l}{\bf{W}}_z^{\rm{T}}{\tilde{\bf{z}}}^{\rm{T}}}$ actually models the correlations among all tokens in $\mathbf{l}_m$ and $\bf{z}$, while ${\bf{W}}_m \in \mathbb{R}^{1\times L}$ is used to aggregate the correlations w.r.t. each token in $\bf{z}$. $\sigma$ denotes the softmax function, which is used to normalize $\bf{w}$ across all $M$ facial components.

Subsequently, we leverage the component weights to refine $\mathbf{z}$ and improve spatial consistency as follows:
\begin{equation}
    {\bf{z}}^{*} = {\bf{z}} + \alpha\sum\limits_{m = 1}^M {{{\bf{w}}_m^{\rm{T}}} \odot \varphi } ({{\bf{l}}_m},{\bf{z}}).
    \label{eq:spatial_enh}
\end{equation}
Here, $\varphi: \mathbb{R}^{L\times D} \times \mathbb{R}^{L'\times D'} \to \mathbb{R}^{L'\times D'}$ is a function that reconstructs ${\bf{z}}$ using the tokens in ${\bf{l}}_m$ as the bases, akin to the cross-attention mechanism for feature interactions \cite{guo2024pulid}, where ${\bf{l}}_m$ serves as both the keys and the values, and ${\bf{z}}$ are considered as the queries. $\odot$ denotes the (broadcastable) element-wise product. $\alpha$ is a hyperparameter controlling the scale of the spatial enhancement. Based on Eqs. (\ref{eq:weight}) and (\ref{eq:spatial_enh}), we can expect that a properly optimized router will assign higher weights to identity-related and discriminative tokens, thereby facilitating effective identity preservation.

As for the local encoder, we adopt the architecture proposed in \cite{yuan2024identity}, which was originally designed for extracting high-frequency patterns, such as contours. However, as the goal of our local encoder is to embed local facial structures, we fine-tune it alongside the local router and the temporal autoregressive module during training our model, rather than keeping it fixed.

\subsection{Temporal Autoregressive Module}
\label{sec:ar}
In addition to the local router for improving frame-wise identity consistency, we introduce the temporal autoregressive module, which aims at cross-frame identity consistency to ensure overall video quality. The core idea of the proposed autoregressive module is to incorporate explicit temporal dependencies into latent tokens to alleviate abrupt changes in appearance and motion.

Figure \ref{fig:tam} demonstrates the architecture of the proposed autoregressive module. This module takes the denoised latent tokens prior to decoding as input, splits them along the frame axis into multiple chunks, and predicts their corresponding biases sequentially for temporal refinement. Moreover, at the beginning of each chunk, the last enhanced frame from the preceding chunk is inserted, thereby forming a ``teacher forcing'' paradigm \cite{williams1989learning} that prevents the enhanced tokens from deviating excessively from the original content. Assume the shape of denoised latent tokens ${\bf{z}}_0$ remains $ L'\times D'$ and the number of chunks is $K$, we denote an arbitrary chunk as ${\bf{c}}_k \in \mathbb{R}^{(1 + \frac{L'}{K}) \times D'}$, $k\in[1,K]$. The temporally enhanced chunk ${\bf{c}}_{k}^{*}$ is obtained as follows:
\begin{equation}
    {\bf{c}}_{k}^{*} = {\bf{c}}_{k} + \beta {\bf{b}}_k, \quad {\bf{b}}_k=\varphi(\psi({\bf{c}}_{k-1}^{*}), {\bf{c}}_{k}),
    \label{eq:ar}
\end{equation}
where ${\bf{b}}_k \in \mathbb{R}^{(1 + \frac{L'}{K}) \times D'}$ denotes the predicted latent biases. $\varphi$ is a cross-attention module with the same architecture used in the local router. $\psi$ is an efficient transformer block that alternately stacks $N$ multi-head self-attention modules and feedforward layers. We employ two techniques within $\psi$ to facilitate temporal context modeling. First, we add the input tokens of $\psi$ with the rotary position embeddings (RoPE) \cite{su2021enhanced} corresponding to their frame indexes to incorporate temporal information. Second, to impose the temporal order constraint on the frames, we employ causal masking \cite{yan2021videogpt} on the outputs of the self-attention modules. Consequently, $\psi$ can capture the temporal contexts in ${\bf{c}}_{k-1}^{*}$ and interact with ${\bf{c}}_{k}$ via $\varphi$, similar to the spatial enhancement paradigm proposed in Eq. (\ref{eq:spatial_enh}). $\beta$ is a hyperparameter for weighting the biases.

\textbf{Discussion.} Figure \ref{fig:tam} summarizes the differences between the proposed autoregressive module with other autoregression methods employed in mainstream generative models. Compared with existing approaches, the advantages of our module are threefold: (i) Our module can fully leverage the superior generation abilities of well-developed diffusion models, compared with methods that merely rely on autoregression \cite{yulanguage,kondratyuk2024videopoet,tian2024visual,li2024controlvar}. (ii) Instead of directly predicting latent tokens, our method predicts biases. This is more stable compared to \cite{ge2022long,han2022show} because the latent tokens initially generated by the baseline LDM serve as a favorable starting point and help to address the accumulation of deviations. (iii) In contrast to hybrid models \cite{ho2022video,chen2024diffusion,ma2024janusflow,ruhe2024rolling}, our module is employed as a plug-in appended to the end of the denoising process. This allows TAM to maintain the original latent space of the base model and improve the temporal consistency efficiently and allows us to train it efficiently (with only one GPU). Besides, temporal correlations in TAM are calculated on the chunk level instead of frame-by-frame, which alleviates drastic latent changes in noisy frames and improves the temporal stability of our synthesized videos.

\subsection{Network Optimization}
\label{sec:loss}
Besides the standard diffusion loss $\mathcal{L}_{diff}$ defined in Eq. (\ref{eq:l_diff}), we propose to use the cross-entropy loss between the output logits of the local router and the ground-truth face segmentation masks for training, namely, 
\begin{equation}
    \mathcal{L}_{route} = - \sum\limits_{m = 1}^M {{{\bf{y}}_m} \odot \log {{\bf{w}}_m}}.
\end{equation}
Here ${{\bf{y}}_m} \in \{0, 1\}^{1\times L'}$ is the ground-truth, where an element equals 1 means that the corresponding token belongs to the $m$-th component, and 0 otherwise. Hence, minimizing $\mathcal{L}_{route}$ encourages the router to recognize local facial structures and facilitate identity preservation.

The total objective function for training our LaVieID framework is given as follows,
\begin{equation}
    \mathcal{L}_\mathrm{total}
    = \lambda_{\mathrm{diff}} \, \mathcal{L}_{\mathrm{diff}}
    \;+\; \lambda_{\mathrm{route}}\,\mathcal{L}_{\mathrm{route}},
\end{equation}
where $\lambda_{\mathrm{diff}}$ and $\lambda_{\mathrm{route}}$ are hyperparameters balancing overall video quality and identity preservation.

\begin{figure*}
    \centering
    \includegraphics[width=1\linewidth]{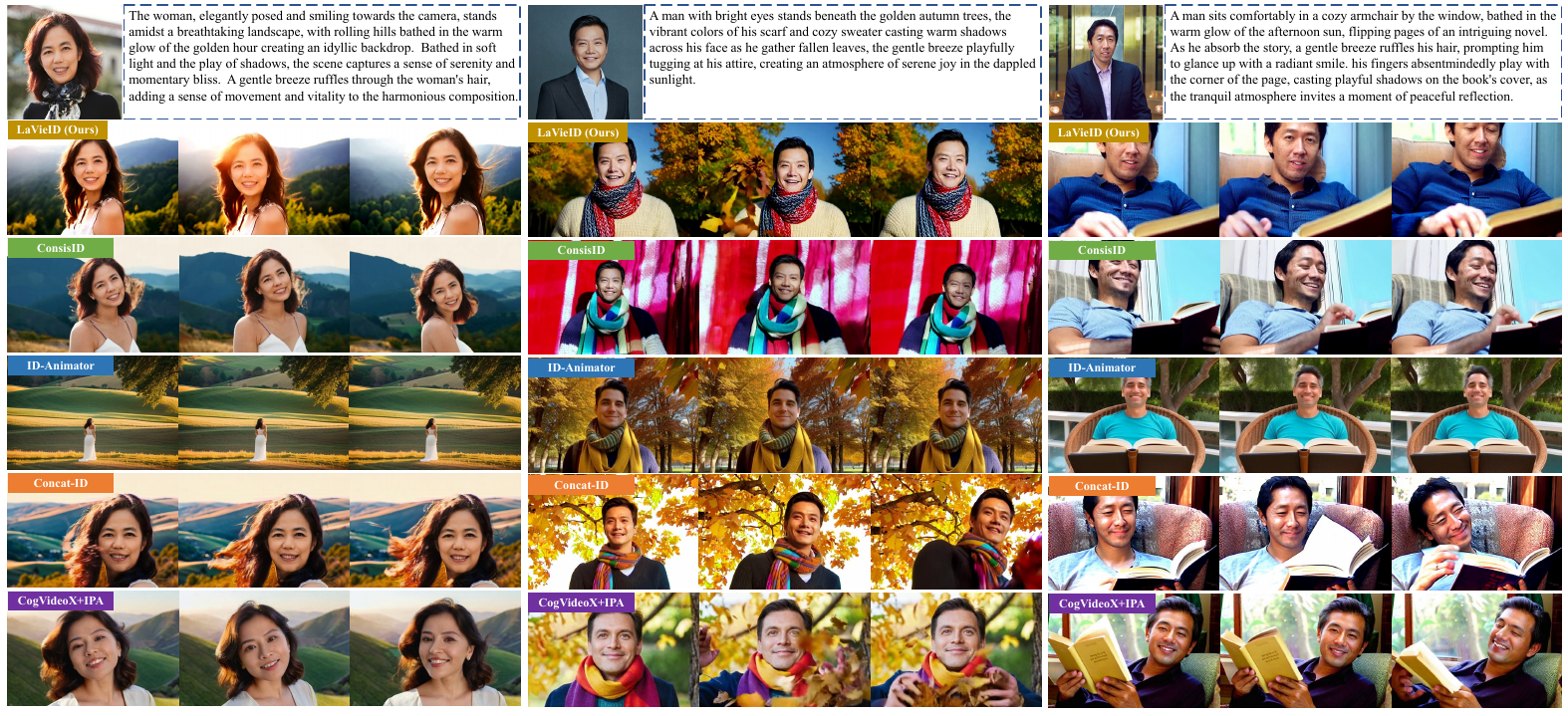}
    \caption{Visual comparisons of the proposed LaVieID against state-of-the-art methods.}
    \label{fig:vis_cmp_sota}
\end{figure*}

\section{Experiment}
\label{sec:exp}
In this section, we validate the proposed LaVieID framework via extensive experiments. Interested readers can refer to our project website for more results and details.

\subsection{Setup}
\label{sec:setup}

\textbf{Implementation details.}
All our experiments are conducted on a single A100 GPU. We use $M=6$ facial component classes, including eyebrows, eyes, mouth, nose, facial skin region, and hair. For local facial token sequences, the number of tokens $L$ is set to 32 and the feature dimension $D$ is set to 2048. The length and feature dimension of latent tokens,$L'$ and $D'$, are set to 17750 and 3072. The inner feature dimension in the local router $D''=2048$ and the transformer block in the temporal autoregressive module contains $N=6$ layers. The number of chunks $K$ is set to 4. The hyperparameters controlling spatial and temporal enhancement, $\alpha$ and $\beta$, are set to 1 and 0.2, respectively. The trade-off weights between $\mathcal{L}_{diff}$ and $\mathcal{L}_{route}$, are set as $\lambda_{\mathrm{diff}} = \lambda_{\mathrm{route}}=1$. We adopt AdamW as the optimizer and train our models with a batch size of 1 and a learning rate of $3 \times 10^{-6}$ for 10K steps, which takes approximately 60 hours. We employ classifier-free guidance (CFG) \cite{ho2021classifier} with random null text prompts at a ratio of 0.1. We use DDIM \cite{songdenoising} for inference with $T=50$ denoising steps and a CFG scale of 6. 

\textbf{Datasets and Metrics.}
We follow the experiment settings of \cite{yuan2024identity} and use its dataset for training and evaluation, including approximately 30K short video clips for training; 30 different subjects disjoint from the training set, with about 5 reference images per subject, and 90 curated prompts for evaluation. We resize all videos to $720 \times 480$ with the frame rate of 8 frames per second and 49 frames per video. To evaluate the proposed method comprehensively, we adopt multiple widely-used metrics, including the subject/background consistency from \cite{huang2024vbench} that measures temporal quality, FID \cite{heusel2017gans} that evaluates the reality of synthesized videos, and FaceSim-Curricular/ArcFace from \cite{deng2019arcface} that calculate the ID similarity between reference images and generated videos.

\begin{table*}[t]
    \centering
    \caption{Quantitative comparisons of the proposed LaVieID framework against state-of-the-art methods.}
    \begin{minipage}{1.0\linewidth}
    \centering
    \begin{tabular}{c|cc|cc|c|cccc}
    \hline
\multirow{2}{*}{Method } &\multirow{2}{*} {\makecell{Subject\\ Consistency} \(\uparrow\)} & \multirow{2}{*} {\makecell{Background \\Consistency} \(\uparrow\) } & \multirow{2}{*} {\makecell{FaceSim\\Curricular} \(\uparrow\) } &\multirow{2}{*} { \makecell{FaceSim\\ArcFace} \(\uparrow\) }&\multirow{2}{*} { \makecell{FID}  \(\downarrow\)}& \multicolumn{4}{c}{Human Evaluation} \\
\cline{7-10}
&&&&&&VQ \(\uparrow\)& TA \(\uparrow\) & DD \(\uparrow\) & IDS \(\uparrow\)   \\
\hline
    ID-animator \cite{he2024id}  &  0.691 & 0.704   & 0.052 & 0.054 & 201.33 & 0.148& 0.154 &0.134& 0.049 \\
    Concat-ID \cite{zhong2025concat} & 0.708 & 0.714  & 0.291 & 0.278 & 204.465 & 0.160& 0.166& 0.150& 0.186 \\
    CogvideoX+IPA \cite{yang2024cogvideox} & 0.743 & 0.768   & 0.105 & 0.074 & \textbf{167.117} & 0.212& 0.188 &0.227 &0.106\\
    ConsisID \cite{yuan2024identity}  & 0.731 & 0.765   &0.412 & 0.389 &  178.447 & 0.131 &0.144& 0.163 & 0.155  \\
        LaVieID (Ours)   &\textbf{ 0.747} & \textbf{0.773}   & \textbf{0.425} &\textbf{0.401} &  174.121 & \textbf{0.350}& \textbf{0.348} & \textbf{0.326}   &\textbf{0.504}\\
    \hline
    \end{tabular} %
    \end{minipage}
    
    \label{tab:qua_cmp_sota}
\end{table*}

\subsection{Comparison with State-of-the-arts}
\label{sec:cmp_sota}

We compare LaVieID with four open-source state-of-the-art IPT2V methods, including ConsisID \cite{yuan2024identity}, ID-animator \cite{he2024id}, CogvideoX+IPA (i.e., CogvideoX \cite{yang2024cogvideox} with IP-Adapter \cite{ye2023ip}), and Concat-ID \cite{zhong2025concat} using the same evaluation protocol.

\begin{figure*}
    \centering
    \includegraphics[width=1\linewidth]{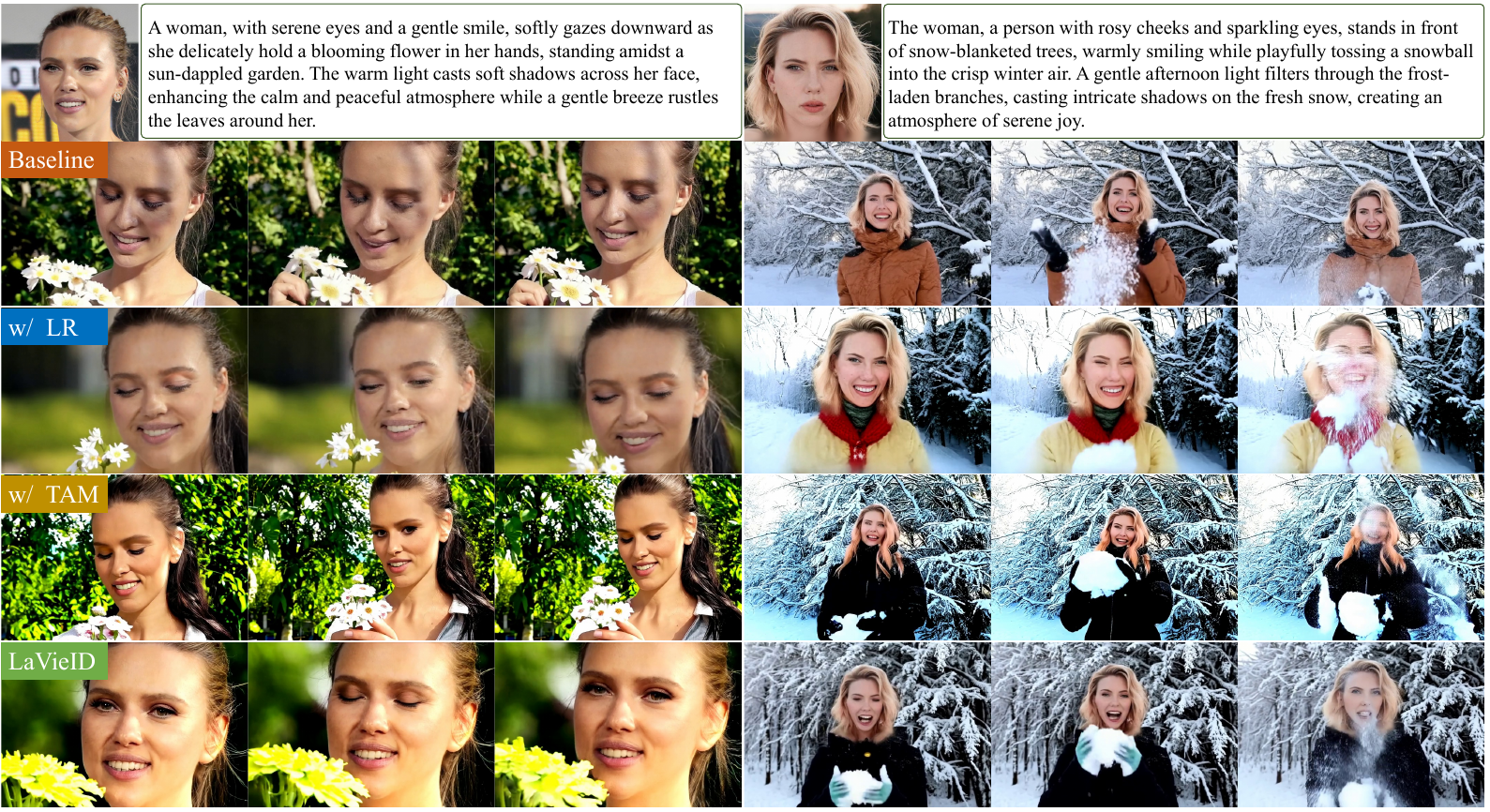}
    \caption{Visual ablation studies on the proposed components.}
    \label{fig:vis_ablation}
\end{figure*}

\textbf{Quantitative analysis.} Table \ref{tab:qua_cmp_sota} summarizes the performance of LaVieID and the baselines. LaVieID outperforms all competitors on identity-centered metrics, including FaceSim-Curricular and FaceSim-ArcFace. It also excels at temporal quality and obtains the highest subject and background consistency values. As for the FID metric, LaVieID is the second best but still slightly worse than CogvideX+IPA (174.121 vs. 167.117). This is because the IPT2V task does not have matched training data and hence can be considered as an out-of-distribution generalization task per se. In this context, CogvideX+IPA is less generalizable, and its results are more likely to follow the distributions of its original large-scale training data. Consequently, its synthesized videos seem more realistic. This can be validated by its FaceSim-Curricular and FaceSim-Arcface scores (0.105 and 0.074), which are much worse than those of LaVieID (0.425 and 0.401). These scores indicate that CogVidex+IPA is less affected by target subjects and is confined to in-domain predictions. On the contrary, LaVieID better balances video quality and identity preservation, obtaining competitive performance on all five metrics.

\begin{table}[t]
    \centering
    \caption{Ablation study on the proposed components.}
    \centering
    \footnotesize
    \begin{tabular}{c|cc|cc|c}
    \hline
Module  & \makecell{Subject\\ Consistency} \(\uparrow\) & \makecell{Background\\ Consistency}  \(\uparrow\) & \makecell{FaceSim\\Curricular} \(\uparrow\) & \makecell{FaceSim\\ArcFace}  \(\uparrow\)& \makecell{FID} \(\downarrow\) \\
\hline

    Baseline  & 0.672 & 0.736  & 0.312 &0.291 &  203.04  \\
    w/ LR  & 0.740 &  0.742  & 0.361 & 0.334 & 178.663  \\
    w/ TAM  & 0.725 & 0.756  & 0.358 & 0.327 & 186.374  \\
    LaVieID  & \textbf{0.755} & \textbf{0.775}  & \textbf{0.428} & \textbf{0.398} & \textbf{168.714}   \\
    \hline
    \end{tabular} %
        
    \label{tab:abaltion_module}
\end{table}

\textbf{Qualitative analysis.} Figure \ref{fig:vis_cmp_sota} provides the visual comparison of video examples generated by different methods. Compared to the baselines, LaVieID produces identity-faithful content of satisfactory quality, even with complex conditions involving diverse poses, motions, expressions, and scenarios. In contrast, ID-animator fails to maintain coherent face representations, and occasionally cannot focus on the subject (e.g., in the first example). ConsisID, while exhibiting relatively stable identity, suffers from poor text alignment in some cases, such as missing the scene description in the second example (e.g., ``golden autumn trees'' and ``leaves''). Moreover, we observe that it changes some facial attributes accidentally. For example, in Figures \ref{fig:teaser} (the second example) and \ref{fig:vis_cmp_sota} (the third example), it mistakenly turns both the rather ``young'' subjects into elders by adding lots of wrinkles. The qualitative examples of CogVideoX+IPA support our opinion that it is ineffective in out-of-distribution generalization and preserving target identity. In summary, the quantitative results of LaVieID align well with its qualitative results, suggesting it is a considerable solution for IPT2V generation.

\subsection{Ablation Studies}
To validate the effectiveness and contributions of each proposed component in LaVieID, we conduct a series of ablation studies. Limited by our computational resources, we select a subset of the test dataset for evaluation in this section. More ablation study results can be found in our project website.

To evaluate the contribution of each key component in LaVieID, we constructed two variants for assessment: one combining the baseline DiT with the local router (denoted as w/ LR) and the other with the temporal autoregressive module (denoted as w/ TAM). The quantitative results of these two variants are reported in Table \ref{tab:abaltion_module}, from which we can observe that both components yield considerable performance gains in all metrics. It should be noted that the objectives of these components differ: the local router aims at spatial identity consistency, whereas the temporal autoregressive module, as its name suggests, is designed for improving temporal identity consistency. Hence, it is reasonable that the baseline w/ LR outperforms its counterpart w/ TAM on identity-centered metrics (FaceSim-Curricular/ArcFace). This can also be supported by a more intuitive comparison based on temporal quality metrics. Specifically, while the baseline w/ LR exhibits higher subject consistency, the baseline w/ TAM achieves superior performance in background consistency. These results suggest that the temporal autoregressive module tends to improve the overall temporal consistency, while the local router focuses more on the subject. Therefore, we can conclude that both components effectively fulfill their intended purposes, and their integration within the LaVieID framework yields the best overall performance.

The visual comparison between the above variants of our method is shown in Figure \ref{fig:vis_ablation}. Compared with the visual examples generated by the baseline w/ LR, those by the baseline w/ TAM exhibit a few identity distortions. For instance, the hairs of both subjects are changed by the latter variant. This is in accordance with our above discussion that the identity consistency benefits more from the local router, while the temporal consistency is improved by the proposed autoregressive module. Another interesting phenomenon is that the baseline w/ TAM tends to generate videos with smaller yet more dynamic subjects, which may also be the reason for its lower performance in subject consistency compared with that of the baseline w/ LR. On the other hand, this also means that the temporal autoregressive module facilitates more diverse and smoother motions. For example, in the left column of Figure \ref{fig:vis_ablation}, the results of the baseline and the variant w/ LR do not exhibit the motion of ``softly gazes downward'', which is achieved by the baseline w/ TAM successfully. This suggests a potential solution to alleviate the limitation of IPT2V models in motion control. That is, the precise generation of subject motions merely by texts is possible, if we can construct the long-range and appropriate temporal correlations between the text descriptions of motions and latent tokens. Finally, LaVieID with the full architecture overcomes the limitations of both the variants and achieves the best visual results.

\subsection{User Study}
\label{sec:user_study}

In the field of video synthesis, whether manually crafted metrics can be fully aligned with human preferences is a controversial and unsolved issue \cite{huang2024vbench,tong2024eyes}. Therefore, we further conduct a user study to provide a subjective evaluation of different methods. Specifically, we assess the quality of generated videos across four dimensions, including visual quality (VQ), text alignment (TA), dynamic degree (DD), and identity similarity (IDS), to provide a comprehensive performance evaluation from the human perspective. We design a questionnaire of 50 questions and invite 30 participants. In each question, we randomly show the synthesized videos of LaVieID and the baselines, and ask the participants to choose the best one based on the above four criteria.

The results of this user study are reported in Table \ref{tab:qua_cmp_sota}, from which we can see that the synthesized videos of LaVieID are favored by most participants across all four criteria. Particularly, more than half of the participants agree that LavieID achieves the best identity-preserving results, which we owe to the proposed local router and the temporal autoregressive module for enhancing spatio-temporal identity consistency.

\section{Conclusion}
This paper introduces a spatio-temporal identity-enhancing framework named LaVieID to complete personalized text-to-video generation. The proposed approach tackles the intrinsic limitations of existing diffusion transformers with two carefully devised components, including a local router and a temporal autoregressive module. The local router dynamically emphasizes fine-grained facial structures to refine latent tokens. The temporal autoregressive module ensures temporal coherence through explicit modeling of long-range dependencies. Extensive quantitative and qualitative experiments and a user study validate that LaVieID significantly surpasses state-of-the-art methods, particularly in terms of identity consistency and visual quality. 

\section*{Acknowledgements}
This work was supported by National Natural Science Foundation of China under Grant No. 62372482 and Shenzhen Science and Technology Program under Grant No. GJHZ20220913142600001.

\bibliographystyle{ACM-Reference-Format}
\balance
\bibliography{sample-base}

\end{document}